\documentclass{article}

\usepackage{arxiv}

\usepackage[utf8]{inputenc} % allow utf-8 input
\usepackage[T1]{fontenc}    % use 8-bit T1 fonts
\usepackage{hyperref}       % hyperlinks
\usepackage{url}            % simple URL typesetting
\usepackage{booktabs}       % professional-quality tables
\usepackage{amsfonts}       % blackboard math symbols
\usepackage{nicefrac}       % compact symbols for 1/2, etc.
\usepackage{microtype}      % microtypography
\usepackage{lipsum}		% Can be removed after putting your text content
\usepackage{graphicx}
\usepackage{natbib}
\usepackage{doi}
\usepackage{amsmath}
\usepackage{algorithm}
\usepackage{algpseudocode}

\usepackage[dvipsnames]{xcolor}

\title{COSMIC: Concurrent Optimization of Structure, Material, and Integrated Control for robotic systems}

\author{ \href{https://orcid.org/0009-0003-4887-6036}{\includegraphics[scale=0.06]{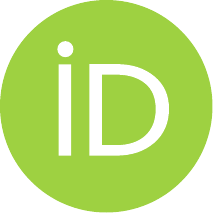}\hspace{1mm}Qinsong Guo} \\
	Dept. of Mechanical Engineering\\
    Carnegie Mellon University\\
    Pittsburgh, PA 15213\\
	\texttt{timg@andrew.cmu.edu} \\
	\And
	\href{https://orcid.org/0000-0002-2777-9718}{\includegraphics[scale=0.06]{orcid.pdf}\hspace{1mm}Liwei Wang}\thanks{Corresponding author} \\
	Dept. of Mechanical Engineering\\
    Carnegie Mellon University\\
    Pittsburgh, PA 15213\\
	\texttt{liweiw@andrew.cmu.edu} 
}

\renewcommand{\shorttitle}{COSMIC}

\hypersetup{
pdftitle={arXiv Robot Co-design},
pdfauthor={Qinsong Guo},
pdfkeywords={co-design, topology optimization, differentiable simulation, machine learning},
}

\begin{document}
\maketitle

\begin{abstract}
	Replicating and surpassing the autonomy of natural organisms remains a long-standing goal in robotics. Yet most robotic systems have their structure, materials, and control designed separately, in sharp contrast to the co-evolution in nature. This separation often leads to suboptimal designs, and we still have a limited understanding of the individual and collective contributions of these design entities. In this work, we propose a gradient-based co-design framework that simultaneously optimizes the topology, material distribution, and control policy of a truss-lattice robot. The framework embeds mixed-type topological and material variables into a continuous design space and integrates a neural network controller within a differentiable simulator, capturing their interactions and enabling efficient gradient calculation via automatic differentiation. Furthermore, we develop a constrained optimization to navigate the highly non-convex design landscape and jointly optimize all design entities. Case studies demonstrate that the proposed framework consistently discovers diverse locomotion strategies that outperform baselines obtained through separated design. The framework is also flexible to accommodate different functional requirements and boundary conditions. Using this framework, we further extract design insights that reveal the individual and collective effects of different entities on robotic performance. The proposed framework provides a computational foundation for the autonomous co-design of robotic systems, capable of reconfiguration, locomotion, and other complex autonomous behaviors.
\end{abstract}

% keywords can be removed

\section{Introduction}

Robots, and machines more broadly, originate from our aspiration to reproduce the autonomy observed in humans and animals, and ultimately surpass natural capabilities \cite{fu_bio-inspired_2014}. To date, substantial progress has been achieved in several key components of robotic systems, i.e., structure \cite{morpho}, materials \cite{kotikian_3d_2018, zhang2024review}, and control \cite{gu_optimization_2025, Aguilar_2016}. However, autonomous robots remain far less adaptive and diverse compared with their natural counterparts \cite{baines_robots_2024}. While advances in soft robotics and responsive materials have expanded the design space \cite{kotikian_3d_2018, zhang2024review}, robot morphologies remain far narrower than the vast diversity found in nature. More fundamentally, structure, materials, and control are often designed separately or sequentially, rather than being co-designed in a tightly coupled manner \cite{morpho, gu_optimization_2025, kobayashi_computational_2024, howard_evolving_2019, schulz_interactive_2017, truss_robot}. For example, a robot structure is typically first specified, after which control strategies are developed to operate the resulting body \cite{guan_design_2023, van_diepen_co-design_2022}. This contrasts with natural systems, where structure, material, and control strategies coevolve to produce highly diverse and specialized organisms \cite{bertossa_morphology_2011}. 

Motivated by this observation, there is growing interest in exploring whether the co-design can lead to performance improvements and increased diversity in artificial robotic systems \cite{matthews_efficient_2023}. This interest in co-design is supported by two considerations. First, from a physical perspective, structure, materials, and control are inherently coupled in system dynamics, producing collective effects that cannot be achieved by each entity independently. Second, from a design perspective, co-design greatly expands the design space\cite{li_generating_2024}, which is more likely to contain a superior optima that is inaccessible to separated design pipelines.

Yet, obtaining an optimal solution via co-design remains challenging in practice. First, the design of different entities often requires variables of different types, leading to a massive and sometimes combinatorial design space. Second, evaluating a robot's dynamic performance is difficult and computationally expensive, involving contacts, friction, large deformations, and nonlinear material behavior\cite{nonlinear, kumar_computational_2018}. This often requires small integration steps over long time horizons, making exhaustive searches prohibitively costly.  Although advanced evolutionary algorithms \cite{li_generating_2024, kriegman_scalable_2020, karacakol_data-driven_2025, strgar_accelerated_2025, strgar_evolution_2024, bhattacharyya_design_2023} and dimensionality-reduction techniques \cite{ schaff2022softrobotslearncrawl, wang_diffusebot_2023} enable more effective search, they still require large populations and many iterations to converge. Third, the physical coupling across design entities makes the underlying optimization problem highly non-convex, even discontinuous; for instance, small changes to structure may lead to large and unpredictable changes in performance. Consequently, comprehensive co-design has mainly been demonstrated for quasi-static systems, such as shape-morphing material networks \cite{wang_autonomous_2025}, kirigami \cite{bordiga_automated_2024}, and granular matter \cite{parsa_gradient-based_2024}, which are computationally cheaper to evaluate with smoother objectives.

 Recent pioneering efforts have begun extending co-design to dynamic systems\cite{MPM, wang2023softzoosoftrobotcodesign, Davy2023magsim, spielberg_advanced_2023, simulator}. In particular, GPU-parallel physics solvers dramatically accelerate evaluations \cite{xue_jax-fem_2023, hu_difftaichi_2020}, while auto-differentiation \cite{project, jia_fenitop_2024} enables efficient calculation of gradients over long temporal horizons for design variables, even in the presence of discontinuous contacts. These advances have enabled a range of frameworks that achieve co-design using gradient-based optimization, drastically reducing the number of simulations compared with exhaustive or heuristic optimizers  \cite{matthews_efficient_2023, spielberg_advanced_2023, sato_computational_2025, yuhn_4d_2023, ma_diffaqua_2021}. Many of these approaches focus on continuum soft robots\cite{wang2023softzoosoftrobotcodesign, spielberg_learning---loop_2019, wang_co-design_2025, bhattacharyya_design_2023}, which allow continuous variations in structures and materials and therefore enable stable optimization\cite{kobayashi_computational_2024, yuhn_4d_2023}. In contrast, co-design of discrete robotic systems remains relatively limited. Their discrete nature introduces mixed-variable design space, sharp discontinuities, and severe nonconvexity in the optimization landscape, making gradient-based co-design more challenging\cite{lumpe_computational_2023}. Yet such systems have many important applications, including legged robots \cite{dinev2022versatile, gao_multi-legged_nodate}, tensegrity structures \cite{caluwaerts2014design, shah_tensegrity_2022}, and bio-hybrid robots \cite{kinjo2024biohybrid, webster-wood_biohybrid_2022}. Therefore, despite these pioneering efforts, there remains a need for comprehensive gradient-based platforms that enable the co-design of structure, materials, and control, particularly for discrete robotic systems. Furthermore, we still lack a clear understanding of the individual and collective contributions of different design entities to the system-level performance \cite{wang_co-design_2025, zhao_topology_2022}. 

\begin{figure*}
    \centering
    \includegraphics[width=0.7\linewidth]{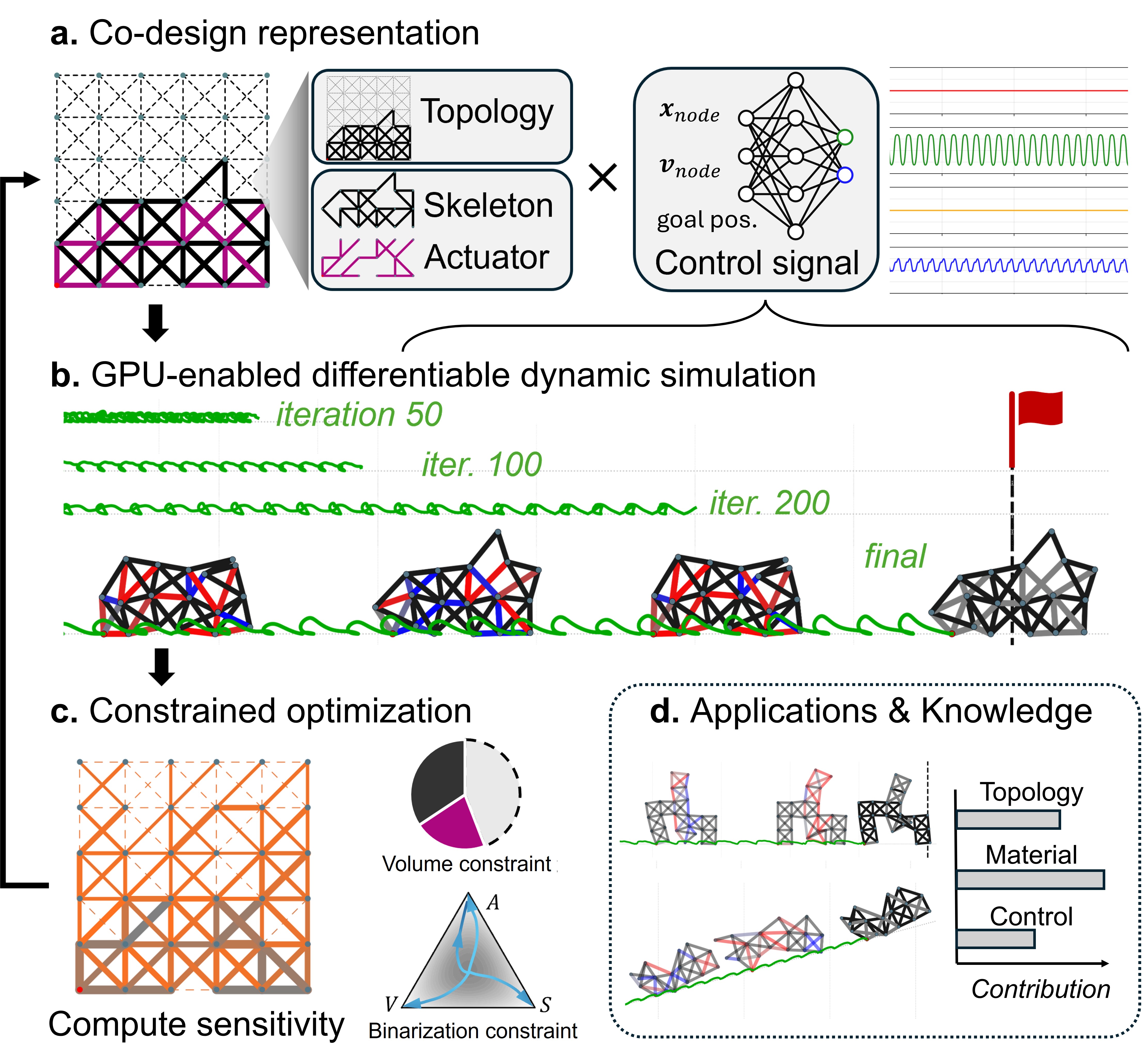}
    \caption{Overview of the dynamic co-design platform}
    \label{fig:overview}
\end{figure*}

In this work, we propose a framework that co-designs the structure, material layout, and control of a discrete truss-lattice robot (Fig.~\ref{fig:overview}). We begin by constructing a unified spatial-temporal design space, merging structure and material selections with control neural network training (Fig. \ref{fig:overview}a). We then develop a GPU-enabled differentiable dynamic simulation to quantify system performance and sensitivity (Fig. \ref{fig:overview}b). We integrate the representation and differentiable evaluation into a constrained optimization to iteratively and concurrently update all design entities (Fig. \ref{fig:overview}c). The framework is demonstrated for a range of dynamic tasks, uncovering valuable knowledge on the effects of co-design entities (Fig. \ref{fig:overview}d).

Our proposed framework has the following contributions:
\begin{itemize}
    \item We introduce a continuous, differentiable design representation that unifies spatial topological, discrete material selection, and spatiotemporal neural control in a discrete robot, enabling the use of efficient gradient-based optimization. 
    \item We develop a GPU-parallel differentiable simulator that captures the physical interdependency among design entities and their environment, efficiently evaluating dynamics in contact-rich scenarios while providing gradient information through automatic differentiation. 
    \item We propose a constrained optimization method that jointly designs structure, materials, and control to automate dynamic robot design. The method incorporates mechanisms to ensure structural integrity and physical feasibility, and automatically generates robotic designs for customized tasks that outperform conventional separate design strategies.
    \item We uncover important design insights by revealing the individual and collective effects of structures, materials, and control on system-level performance.
\end{itemize}

The remainder of this paper is structured as follows. We first detail the proposed co-design framework (Section \ref{Section2}), and then demonstrate its effectiveness against baselines (Section \ref{section31}). Next, we employ an ablation study to investigate the individual and collective contributions of design entities (Section \ref{section32}). Finally, we demonstrate the generalizability of the framework across different objectives, physical environments, and design spaces (Section \ref{section33}). Conclusions are drawn in Section \ref{conclusion}.

\section{Method}
\label{Section2}

In this section, we develop a general co-design framework for dynamic robotic systems. We consider a lattice robot composed of truss elements, where each truss can function either as a passive skeleton or as an externally controlled actuator. Our goal is to co-design the topology, the spatial distribution of material types, and the temporal control signals for each actuator to maximize dynamic performance. To achieve this, we extend topology optimization to integrate a mixed-variable design representation, a differentiable dynamic simulation for design evaluation, and a constrained optimization for design synthesis.

\subsection{Design Representation}

\subsubsection{Topology and Material Layout}
\begin{figure}
    \centering
    \includegraphics[width=0.75\linewidth]{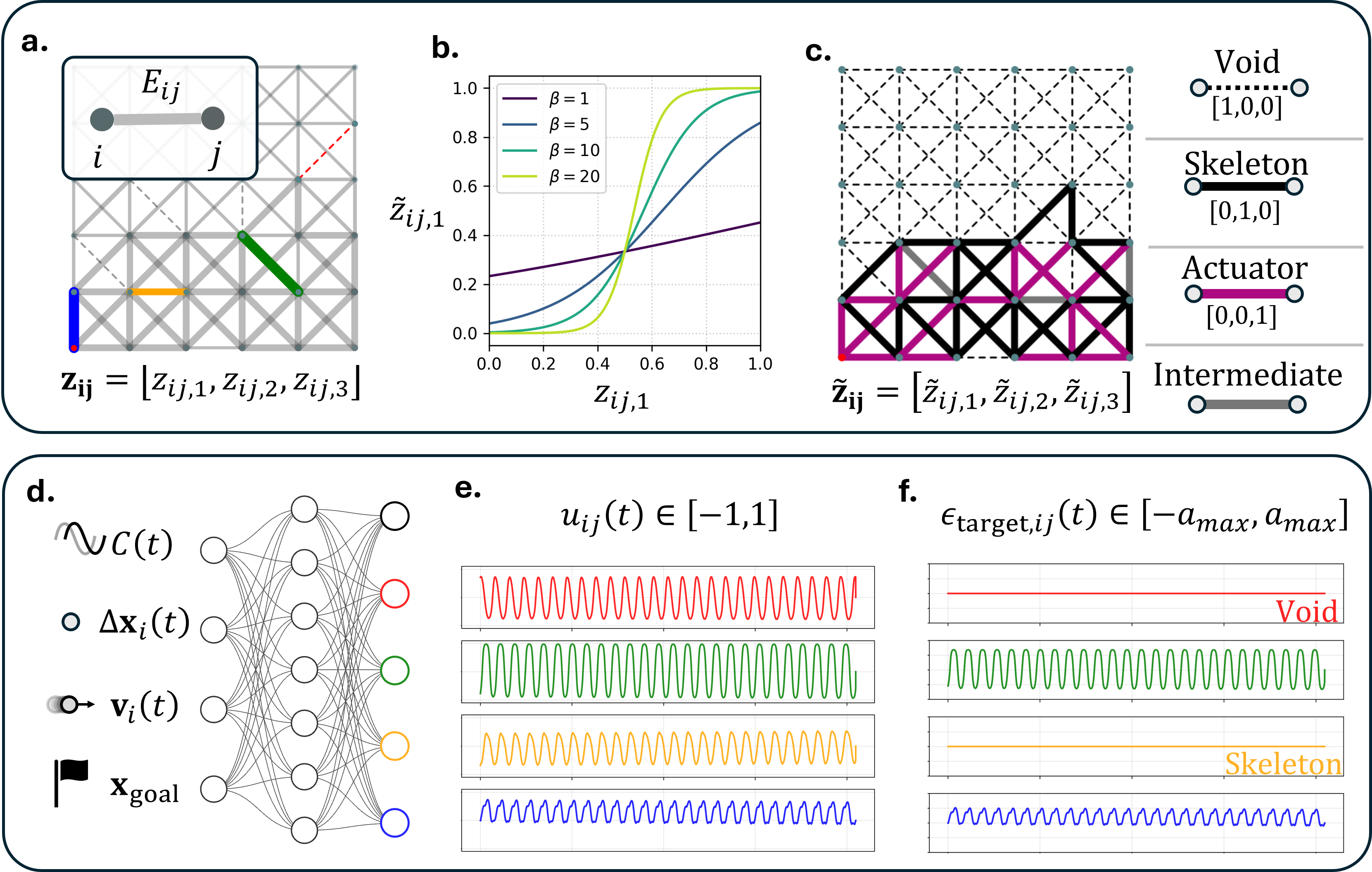}
    \caption{ Design representation: (a) Each edge $E_{ij}$ is associated with a continuous, relaxed design vector $\mathbf{z}_{ij}$. (b) Through a softmax projection scaled by $\beta$, (c) we project it to a one-hot-encoded material state vector $\mathbf{\tilde z}_{ij}$, that dictates the mix ratio between discrete material types. (d) The Multi-Layer Perceptron controller receives synchronized clock signals from a Central Pattern Generator, the instantaneous nodal positional offsets and velocity, and a 2D goal vector to (e) output a control signal for each edge $u_{ij}(t) \in [-1, 1]$. (f) The control signal is scaled by edge material properties, including $a_{\max}$ and $\tilde{z}_{ij,3}$ to yield the target strain for each edge at time $t$, $\epsilon_{\text{target},{ij}}(t)$. Notice how only edges with the actuator material state have non-zero target strain. } 
    \label{fig:representation}
\end{figure}
\quad Consider a lattice robot composed of truss elements connected at mass nodes, as shown in Fig. \ref{fig:representation}. The structure can be represented as a graph with $N_m$ nodes and $N_e$ edges $E_{ij}$, where each edge connects node $i$ and node $j$. The set of all connected node pairs is denoted by $E = \{(i,j)\}$. Each edge can take one of three possible states: void, skeleton, and actuator. To represent this edge-wise design choice, we associate each edge $E_{ij}$ with a three-dimensional one-hot vector $\mathbf{z}_{ij} = [z_{ij,1}, z_{ij,2}, z_{ij,3}]^\top \in \{0,1\}^3$, where the three entries correspond to the states void, skeleton, and actuator, respectively (Figure \ref{fig:representation}c). When an edge is assigned a specific state, the corresponding entry equals one while the others are zero. This discrete representation poses challenges for gradient-based optimization. Therefore, we relax the design variable to a continuous vector $\mathbf{z}_{ij} \in [0,1]^3$ (Figure \ref{fig:representation}a) and project it to a normalized material state vector $\mathbf{\tilde{z}}_{ij}$ using a scaled softmax function that approximates the original one-hot encoding:
\begin{equation}\label{eq:softmax_general}
\tilde{z}_{ij,k} = \frac{\exp(\beta z_{ij,k})}{\sum_{l=1}^{3} \exp(\beta z_{ij,l})},
\end{equation}
% To evaluate the structural performance, the design logits are projected into a material probability vector $\mathbf{p}_s \in [0,1]^N$ via a temperature-scaled softmax function:
% \begin{equation}\label{eq:softmax_general}
%     p_{s,c} = \frac{\exp(\beta z_{s,c})}{\sum_{k=1}^N \exp(\beta z_{s,k})}
% \end{equation}
where $\beta$ is a scaling factor to control the projection sharpness, as shown in Figure \ref{fig:representation}b. To further promote binarization, we introduce an explicit binarization constraint in the optimization problem (Section.~\ref {subsec:design_opt}), which can adaptively penalize intermediate values to drive the design variables toward one-hot encodings that correspond to discrete material states.

To interface with the physical simulation, each material state is associated with $H$ mechanical properties, stored in a material property library matrix $\Psi \in \mathbb{R}^{N_e \times H}$. The effective property vector $\mathbf{p}_{ij} \in \mathbb{R}^H$ for truss $E_{ij}$ is computed via matrix multiplication:
\begin{equation}\label{eq:mat_prop}
    \mathbf{p}_{ij} = \Psi^\top \mathbf{z}_{ij},
\end{equation}
where the entries of $\mathbf{p}_{ij}$ dictate the interpolated mechanical properties (e.g., stiffness, density) to be used in simulation.

In this study, we consider $H=3$ properties in the property matrix $\Psi$, including the stiffness ($\text{k}$), linear density ($\rho$), and maximum actuation strain ($a_{\max}$). Each material state is characterized by a distinct combination of these properties:

\begin{itemize}
    \item \textbf{Void ($V$):} has ultra-low stiffness, ultra-low linear density, and zero actuation capability. Trusses assigned to this state effectively act as a void to allow topological changes.
    \item \textbf{Skeleton ($S$):} has high stiffness, medium density, and zero actuation capability. It mainly provides the structural support of the robot. Its high stiffness allows it to passively resist deformation, upholding the structural integrity vital for the stability of a discrete lattice.
    \item \textbf{Actuator ($A$):} has a stiffness an order of magnitude lower than the skeleton, and a linear density three times as high (mirroring the heavy weight of real-world electromechanical components). While compliant and heavy, it possesses high actuation capabilities.
\end{itemize}

This property scheme reflects the inherent trade-off observed in natural and artificial systems, where a truss cannot simultaneously possess high stiffness and high actuation capacity. Therefore, to maximize performance within a limited materials constraint, it requires a highly strategic, synergistic allocation of passive skeleton and active actuation.

Finally, to accelerate the simulation, we lump the mass of the trusses at the connecting nodes. The mass $m_i$ of node $i$ is computed as:
\begin{equation}\label{eq:mass}
    m_i = m_\epsilon + \frac{1}{2} \sum_{j \in \mathcal{N}(i)} l_{ij} \sum_{k =1}^3 \tilde{z}_{ij,k} \rho_k,
\end{equation}
where $\mathcal{N}(i)$ denotes the set of nodes connected to node $i$, $l_{ij}$ is the resting length of edge $E_{ij}$, and $\rho_k$ is the linear density associated with material state $k$. To maintain numerical stability and prevent singularities during the physics simulation, every node is initialized with a negligible baseline mass $m_\epsilon$. Consequently, nodes connected exclusively to void trusses act as negligible masses, naturally eliminating themselves from the system dynamics. The head node $i=0$ is instantiated with a larger weight, acting as a payload to be carried forth by the robot. 

\subsubsection{Actuation Control}
\quad The dynamic actuation of the robot is governed by a Multi-Layer Perceptron (MLP) neural network, as shown in Figure \ref{fig:representation}d. The MLP processes the instantaneous kinematic state of the robot alongside a synchronized clock signal to output a target actuation length for each truss. Specifically, the network input vector $\mathbf{I}(t)$ at time $t$ consists of Central Pattern Generator (CPG) signals \cite{ijspeert2008central}, nodal positional offsets, nodal velocity vectors, and a goal position vector. The CPG signals $\mathbf{C}(t) \in \mathbb{R}^{N_{\text{CPG}}}$ are modeled as phase-shifted sine waves to ensure all active trusses actuate in a synchronized, periodic manner:
\begin{equation}\label{eq:cpg}
    C_j(t) = \sin\left(\omega t + \frac{2\pi j}{N_{\text{CPG}}}\right), \quad j \in \{0, \dots, N_{\text{CPG}}-1\}
\end{equation}
Let the position vector of node $i$ be defined as $\mathbf{x}_i(t) = [x_i(t), y_i(t)]^\top$. The positional offsets are thus $\Delta \mathbf{x}_i(t)$, which provide spatial awareness by tracking the displacement of each node $i \in \mathcal{M}$ relative to the geometric center of the robot:
\begin{equation}\label{eq:offset}
\Delta \mathbf{x}_i(t) = \mathbf{x}_i(t) - \frac{1}{N_m} \sum_{j=1}^{N_m} \mathbf{x}_j(t)
\end{equation}
Additionally, we define a 2D goal vector $\mathbf{x}_{\text{goal}}\in\mathbb{R}^2$ that specifies the desired target coordinates for the robot's locomotion. 

By concatenating these features, we obtain the complete input vector:
\begin{equation}\label{eq:nn_input}
    \mathbf{I}(t) = \left[ \mathbf{C}(t)^\top, \mathbf{x}_{\text{goal}}, \Delta \mathbf{x}_1(t)^\top, \dots, \Delta \mathbf{x}_{N_m}(t)^\top, \mathbf{v}_1(t)^\top, \dots, \mathbf{v}_{N_m}(t)^\top \right]^\top
\end{equation}
While the CPG provides the fundamental gait frequency, positional and velocity information allows the control policy to be modulated by the gait cycle's current stage, helping it react to environmental perturbations by sensing velocity changes. Further, the inclusion of the goal vector enabled active steering toward different spatial targets. 

The MLP comprises a single hidden layer with a bias vector, followed by an output layer of dimension $N_e$, i.e., the total number of edges. We utilize a hyperbolic tangent ($\tanh$) activation function to bound the network outputs $u_{ij}(t) \in [-1, 1]$, where $1$ denotes maximum extension and $-1$ denotes maximum contraction (Figure \ref{fig:representation}e). While the control neural network outputs signals for all edges, only those edges assigned to the actuator state physically respond. To enforce this, we modulate the raw MLP outputs based on edge-wise material state vector to obtain the effective target strain $\epsilon_{\text{target},{ij}}(t)$ as:
\begin{equation}\label{eq:strain}
    \epsilon_{\text{target},{ij}}(t) = \tilde{z}_{ij,3} \cdot a_{\max} \cdot u_{ij}(t),
\end{equation}
where $\tilde{z}_{ij,3}$ is the assigned ratio of Actuator for truss $E_{ij}$.
This way, only trusses in actuator states will receive a target strain (Figure \ref{fig:representation}d). 
\subsection{Design Evaluation}
\label{sec:evaluation}

\begin{figure}
    \centering
    \includegraphics[width=0.9\linewidth]{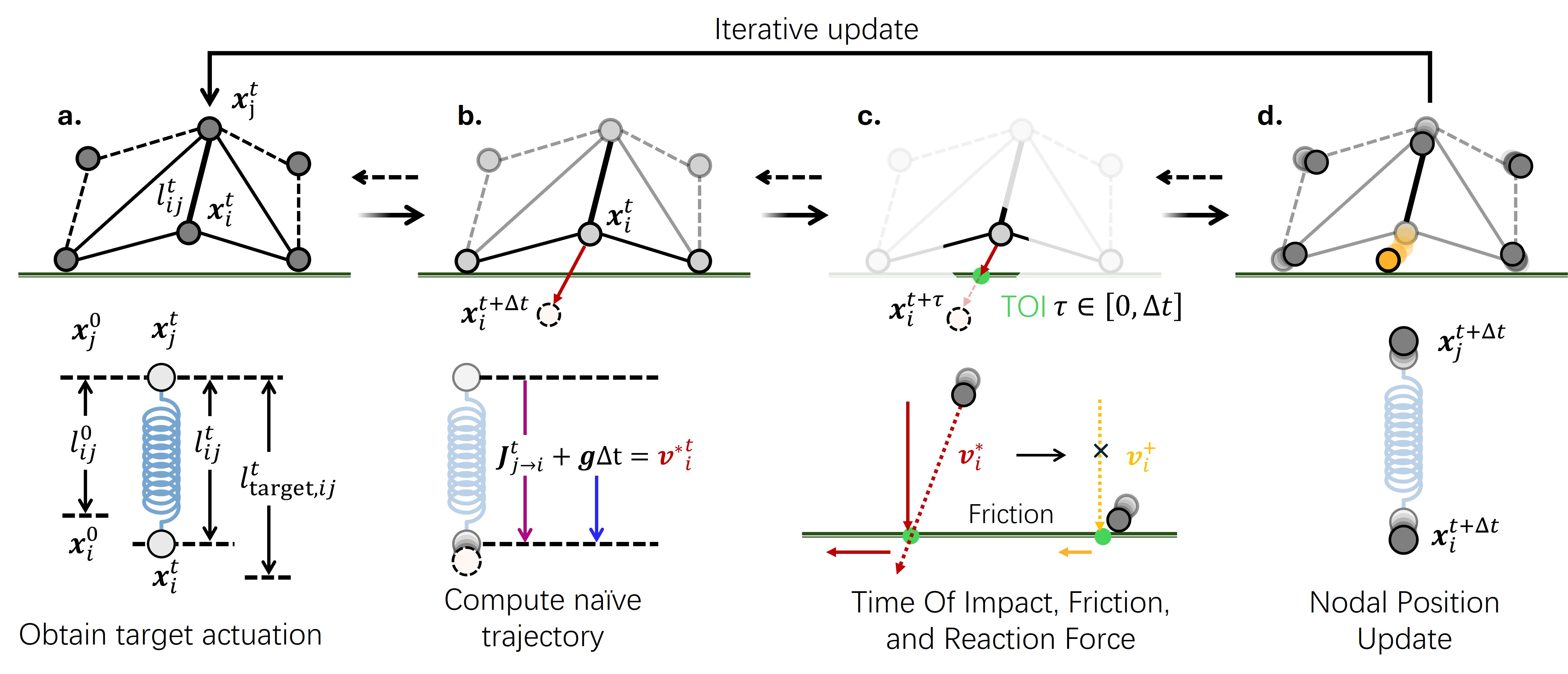}
    \caption{Illustration of the differentiable simulation. At integration time $t$, (a) a generalized Hooke's law is used to determine the force produced by edge $E_{ij}$ from its target actuation length $l_{\text{target},ij}(t)$ and current length $l_{ij}(t)$. (b) The pre-collision velocity vector $\mathbf{v}^{*}_i(t)$ is calculated from the resultant aggregated force on node $i$ to update the nodal position $\mathbf{x}_i(t+\Delta t)$. (c) The position and velocity of nodes that penetrate the ground are modified using Time Of Impact (TOI) techniques. At TOI, we compute a post-collision velocity vector $\mathbf{v}^{+}_i(t)$ by applying frictional damping to the pre-collision tangential component, and eliminating the perpendicular component. (d) Final nodal positions are obtained, advancing the simulation to the next step $t+\Delta t$, completing a forward simulation. Gradients are computed by propagating backwards in the computational graph (dashed arrows).}
    \label{fig:simulation}
\end{figure}
We further couple the proposed design representation with differentiable simulators to evaluate the system performance. We assume truss forces are governed by a generalized Hooke's law, with stiffness and actuation properties computed in Eq.~\ref{eq:mat_prop}. For example, the effective stiffness $\text{k}_{ij}$ of truss $E_{ij}$ is computed as the dot product of its material state vector and the stiffness values associated with each state: $\text{k}_{ij} = \tilde{z}_{ij,1} \text{k}_1 + \tilde{z}_{ij,2} \text{k}_2 + \tilde{z}_{ij,3} \text{k}_3$.

The target actuation length of the truss is determined dynamically by the target strain $\epsilon_{\text{target},ij}(t)$ generated by the MLP controller:
\begin{equation}\label{eq:length}
    l_{\text{target},ij}(t) = l_{0,ij} \left[ 1 + \epsilon_{\text{target},ij}(t) \right],
\end{equation}
where $l_{0,ij}$ is the original length of the truss without actuation, as shown in Figure \ref{fig:simulation}a. The internal scalar forces can then be calculated by Hooke's law:
\begin{equation}\label{eq:force}
    F_{ij}(t) = k_{ij} \left[l_{ij}(t) - l_{\text{target},ij}(t) \right]
\end{equation}
Intuitively, a larger commanded target strain magnitude results in a greater difference between the current length and the target length, thereby producing a larger internal force in the truss. To simulate the system dynamics, we integrate the force over the simulation time step $\Delta t$ and transfer it to the connecting nodes $i$ and $j$ as a nodal impulse. Specifically, for the nodal force imposed on node $i$ from the truss $E_{ij}$:
\begin{equation}\label{eq:impulse}
    \mathbf{J}_{j \rightarrow i}(t) = F_{ij}(t) \cdot \Delta t \cdot \mathbf{u}_{ij}(t),
\end{equation}
where $\mathbf{u}_{ij}(t) = \frac{\mathbf{x}_j(t) - \mathbf{x}_i(t)}{l_{ij}(t)}$ is the unit directional vector pointing from node $i$ to node $j$. We then aggregated all nodal impulses to update the nodal position. To simulate atmospheric drag and internal energy dissipation, we also apply an exponential frictional damping decay to the nodal velocities. The nodal location and velocity $\mathbf{v}_i$ for node $i$ is updated via symplectic Euler integration:
% \begin{equation}\label{eq:damping}
%     \mathbf{v}_i(t+\Delta t) = \exp(-\gamma \Delta t) \mathbf{v}_i(t) + \frac{1}{m_i} \sum_{j \in \mathcal{N}(i)} \mathbf{J}_{j \rightarrow i}(t) + \mathbf{g} \Delta t
% \end{equation}
\begin{equation}\label{eq:damping}
\begin{aligned}
\mathbf{v}_i(t+\Delta t) &= \exp(-\gamma \Delta t) \mathbf{v}_i(t)
+ \frac{1}{m_i} \sum_{j \in \mathcal{N}(i)} \mathbf{J}_{j \rightarrow i}(t)
+ \mathbf{g} \Delta t,\\
&\mathbf{x}_i(t+\Delta t) = \mathbf{x}_i(t) + \mathbf{v}_i(t+\Delta t)\Delta t,
\end{aligned}
\end{equation}
where $\gamma$ is the damping coefficient and $\mathbf{g}$ is the gravity vector (Figure \ref{fig:simulation}b).

To handle the contact with the ground, we use a high-fidelity Time of Impact (TOI) scheme to prevent interpenetration and reduce gradient discontinuity \cite{hu_difftaichi_2020}. Upon ground contact detection, the exact TOI $\tau \in [0, \Delta t]$ is linearly interpolated from the pre-collision trajectory (Figure \ref{fig:simulation}c). When considering an infinite friction scenario, a zero-velocity boundary condition is applied at the contact point. When Coulomb friction is considered, the collision is modeled as perfectly inelastic in the normal (vertical) direction, while the tangential (horizontal) velocity is penalized by the coefficient of dynamic friction $\mu$:
\begin{equation}\label{coulomb}
\begin{aligned}
v_{i,y}^+(t) &= 0 \\
v_{i,x}^+(t) &= \begin{cases}
0, & \text{if infinite friction}\\
v_{i,x}^*(t) - \mu |v_{i,y}^*(t)| \text{sgn}(v_{i,x}^*(t)), & \text{if Coulomb}
\end{cases}
\end{aligned}
\end{equation}
where $\mathbf{v}_i^*(t)$ and $\mathbf{v}_i^+(t)$ denote the pre-collision and post-collision velocities, respectively, and $\mu$ is the coefficient of dynamic friction. Note that this velocity modification will be imposed before updating the nodal position in Eq.~\ref{eq:damping}. 

% The overall robot performance is assessed by a displacement loss term, which is the negative horizontal distance traveled by the payload head node (index $0$) at the final rollout timestep $T$:
% \begin{equation}\label{eq:loss_disp}
% \mathcal{L}_{\text{disp}} = -x_0(T)
% \end{equation}

By iteratively applying this update rule, we obtain the full motion trajectory of the robot through forward simulation (Figure \ref{fig:simulation}). The resulting procedure also defines a forward computational graph, which allows automatic differentiation to backpropagate the sensitivity of the nodal displacement at the final state through all integration time steps to the design variables and weights of control networks. This yields the gradients required to update the design variables and control weights.

\subsection{Design Optimization}
\label{subsec:design_opt}
While the proposed framework is flexible enough to accommodate various functional objectives, we focus on forward locomotion in this study where the goal is to maximize the traveled distance. The co-design problem is formulated as a joint optimization over the structural state variables $\mathbf{Z} = \{\mathbf{z}_{ij}\}$ and the neural network controller parameters $\boldsymbol{\theta}$. The objective is defined as the negative forward displacement of the payload head node (index $0$) at the final rollout timestep $T$:
\begin{equation}
\min_{\mathbf{Z}, \boldsymbol{\theta}} \; \mathcal{L}_{\text{disp}}(\mathbf{Z}, \boldsymbol{\theta}) = -x_0(T)
\end{equation}

To obtain feasible solutions, we introduce additional constraints to the optimization problem and incorporate structural stability checks when solving the problem, as described in the following.

\subsubsection{Constraints}
\begin{figure}
\centering
\includegraphics[width=0.6\linewidth]{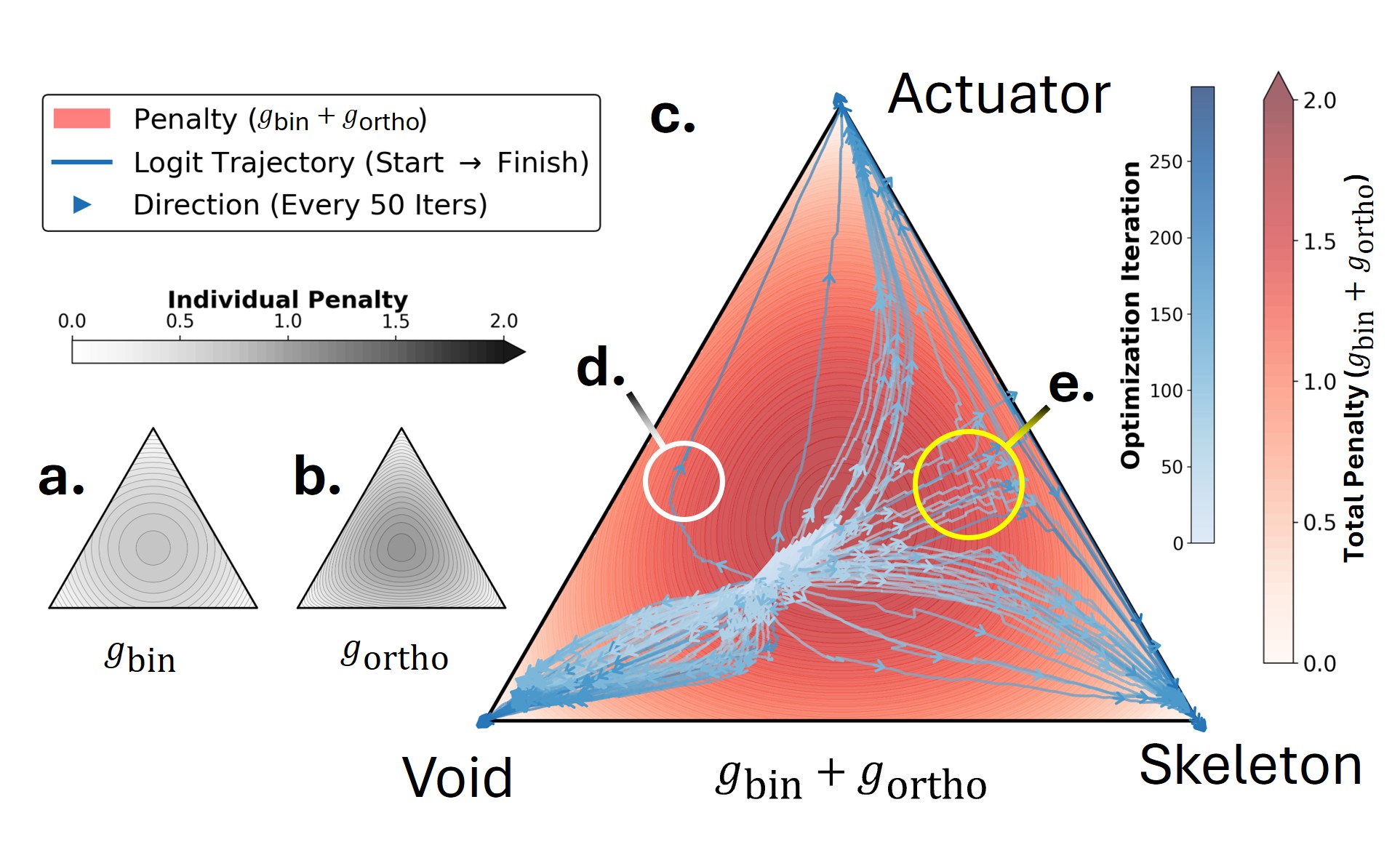}
\caption{Penalty fields of (a) the global binarization constraint, (b) the pairwise orthogonal constraint, and (c) the combined penalty constraint with the optimization trajectory of design vectors throughout iterations. The three corners of each ternary plot represent discrete states, while the interior indicates intermediate interpolations between them. (d) Circled in white: an edge that struggles to decide between actuator and void, but converges towards actuator, facilitated by the iterative attraction phase. (e) Circled in yellow: zigzag design vector trajectories due to the competition between displacement and stability objectives.}
\label{fig:manifold}
\end{figure}

\quad We impose two main categories of constraints: volume fraction constraints and binarization constraints. First, we enforce volume fraction constraints on different material states. The total solid volume fraction $V$ (including both actuator and skeleton) and the actuator volume fraction $V_{\text{act}}$ are computed as
\begin{equation}\label{eq:volume}
V = \frac{1}{N_e} \sum_{(i,j)\in E} (p_{ij,2} + p_{ij,3}),
\quad
V_{\text{act}} = \frac{1}{N_e} \sum_{(i,j)\in E} p_{ij,3}
\end{equation}
These scalar volume fractions are required to satisfy the prescribed bounds $(V^{\max}, V^{\min})$ and $(V_{\text{act}}^{\max}, V_{\text{act}}^{\min})$, respectively. To provide mobility for \textit{intermediate} material states ($\tilde z_{ij,k} \notin \{0,1\}$) that struggle to satisfy competing objectives, we progressively relax these bounds up to a small tolerance $\Delta_V$. This way, task performance will be prioritized over a strictly satisfied volume bound, which may damage physical stability. Consequently, the effective bounds become $V^{\max*} = V^{\max} + \Delta_V$, $V^{\min*} = V^{\min} - \Delta_V$, and similarly for the actuator volume. 

Second, we impose binarization constraints on the material state vectors to ensure they eventually achieve a fully discretized state, i.e., $\tilde{z}_{ij,k} \in \{0,1\}$ (Fig. \ref{fig:manifold}). This is accomplished through a global binarization penalty $g_{\text{bin}}$ that drives the $L_2$ norm of the material state vector toward 1 \cite{yuhn_4d_2023}, and a pairwise orthogonal penalty $g_{\text{ortho}}$ that minimizes the overlap between entries corresponding to any two distinct material types:
\begin{equation}\label{eq:loss_bin}
    g_{\text{bin}} =  \frac{1}{N_e} \sum_{(i,j)\in E} \left( 1 - \sqrt{\sum_{k=1}^3 \tilde{z}_{ij,k}^2} \right),
\end{equation}
\begin{equation}\label{eq:loss_ortho}
    g_{\text{ortho}} =  \frac{1}{N_e} \sum_{(i,j)\in E} \sum_{k \neq l} \frac{\tilde{z}_{ij,k} \tilde{z}_{ij,l}}{0.5(\tilde{z}_{ij,k} + \tilde{z}_{ij,l}) + \delta},
\end{equation}
where $\delta$ is a small value to avoid singularity. These penalty values should be controlled within a given acceptable range.

Note that in standard density-based topology optimization, methods like Solid Isotropic Material with Penalization (SIMP) are frequently used to penalize the stiffness of intermediate probabilities, driving the optimizer toward discrete solutions \cite{bendsoe_optimal_1989}. However, due to the extreme non-convexity of the optimization landscape for discrete truss robots, such penalization can induce significant instability in the system dynamics, as observed in our preliminary tests. This motivates our use of the binarization constraints described above, which adaptively guide the solution toward discrete states over time.
% \begin{figure}
%     \centering
%     \includegraphics[width=1\linewidth]{fig/sequence.jpg}
%     \caption{Optimization history of the co-design process corresponding to the design in Figure.~\ref{fig:init}b. (a–d) Intermediate designs at iterations 20, 70, 110, and 200, respectively, showing the evolution of the structure. (d) Structural fine-tuning during the design vector attraction stage. (e) Fully binarized and frozen structural layout in the control-only design stage, where only the MLP controller weights are fine-tuned.}
%     \label{fig:sequence}
% \end{figure}

\subsubsection{Optimization Problem and Solution}

\quad With the above constraints, our goal is to find the optimal combination of the material design vectors $\mathbf{Z} = \{\mathbf{z}_{ij}\}$ and the neural network weights $\boldsymbol{\theta}$ that maximizes the locomotion distance. This co-design problem can be formulated as a constrained optimization:

% The primary \textit{performance pass} evaluates the continuous material distribution to update both the controller and the structural design; this is interleaved with a \textit{stability pass} that operates on the highly discretized proxy geometry $\mathbf{p}^*$. The multi-objective constrained optimization problem is defined as:
\begin{equation}\label{eq:opt_problem}
\begin{aligned}
\min_{\mathbf{Z}, \boldsymbol{\theta}} \quad & \mathcal{L}_{\text{disp}}(\mathbf{Z}, \boldsymbol{\theta}) \\
% \quad &\text{(Performance Pass)} \\
% \min_{\mathbf{Z}} \quad & \mathcal{L}_{\text{stab}}(\mathbf{p}^*(\mathbf{Z})) \quad &\text{(Stability Pass)} \\
\text{subject to}\quad& V^{\min*} \le V(\mathbf{Z}) \le V^{\max*}, \\
& V_{\text{act}}^{\min*} \le V_{\text{act}}(\mathbf{Z}) \le V_{\text{act}}^{\max*}, \\
& g_{\text{bin}}(\mathbf{Z}) \le g_{\text{bin}}^*, \\
& g_{\text{ortho}}(\mathbf{Z}) \le g_{\text{ortho}}^*, \\
% & \text{Interpolation Eqs. (\ref{eq:softmax_general}), (\ref{eq:mat_prop}), (\ref{eq:stab_softmax}),} \\
% & \text{Governing Physics Eqs. (\ref{eq:length}), (\ref{eq:force}), (\ref{eq:damping}), (\ref{coulomb}).}
\end{aligned}
\end{equation}
where $g_{\text{bin}}^*$ and $g_{\text{ortho}}^*$ represent the acceptable threshold for the binarization and pairwise orthogonal constraints, respectively.

We solve this constrained problem via iterative gradient-based optimization. To promote the initial exploration of vast design possibilities, binarization constraints are enforced through the Augmented Lagrangian (AL) method \cite{yuhn_4d_2023, fortin2000augmented}. Unlike volume fraction constraints, which are strictly enforced at all times, binarization constraints are allowed to be occasionally violated to increase the exploration of discrete choices and open up potential pathways to better solutions. The AL method provides a rigorous mechanism to adaptively adjust the penalization strength to achieve this. Empirically, we find this approach yields better performance compared with strictly enforcing the constraints. The objective function with AL constraints is defined as:
\begin{equation}
    \mathcal{L}(\mathbf{Z}, \boldsymbol{\lambda}, \boldsymbol{\tau}) = \mathcal{L}_{\text{disp}} + \sum_{k=1}^4 \left( -\lambda_k v_k + \frac{1}{2} \tau_k v_k^2 \right),
\end{equation}
where $v_k$ denotes the violation magnitude of the $k$-th binarization constraint (one global $L_2$ penalty and three pairwise orthogonal penalties). The Lagrangian multipliers $\boldsymbol{\lambda}$ are updated dynamically at each iteration according to $\lambda_k \leftarrow \lambda_k - \tau_k v_k$. To prevent over-penalization, the penalty weight $\tau_k$ is increased by a factor of $a$ if the constraint violation $v_k$ increases between iterations, and decreased by $1/a$ if the violation decreases.

All material design vectors $\mathbf{Z}$ are initialized uniformly and updated iteratively via the Method of Moving Asymptotes (MMA). The MMA optimizer takes in the augmented objective $\mathcal{L}(\mathbf{Z}, \boldsymbol{\lambda}, \boldsymbol{\tau})$, the volume fraction constraints, and gradients from the differentiable simulation. % Inequality volume constraints are passed directly into the MMA optimizer to be more strictly satisfied. Conversely, the AL method adaptively ramps up penalization strength throughout the optimization iterations, enabling a smooth and continuous transition from exploration to exploitation. 
The MLP controller weights $\boldsymbol{\theta}$ are initialized using Xavier Initialization and updated using the Adam optimizer, which represents the state-of-the-art for gradient-based optimization over large parameter sets in neural networks. Material design vectors and MLP weights are updated alternately at a 3:2 ratio, based on our preliminary experiments, to enable a balanced update between robot morphology and control, supporting optimal co-evolution of the configuration. 

% Despite the binarization constraint penalization, we find in practice that it is still possible for some edges to get trapped into intermediate materials that form "super materials" with considerable actuator force and skeleton stiffness, which are mechanically advantageous in simulation but physically infeasible.
Despite the binarization constraint penalization, in practice, some edges can still become trapped in intermediate material states in the optimization (Fig. \ref{fig:manifold}e). In these cases, the dynamic system relies on a delicate structural integrity and balance that is maintained by these physically infeasible intermediate materials. Once these materials are removed, the delicate balance may be disrupted, potentially causing collapse of the structure or disjoint geometry, which can drastically alter the system dynamics. Consequently, performance losses evaluated on intermediate designs rarely reflect the true performance of the final discrete robot.

To reliably probe the structural integrity of the final discrete robot, we propose to implement an auxiliary stability evaluation along with the ordinary objective function evaluation. This pass implements a globally centered softmax function to project the continuous design variables $\mathbf{z}$ into a highly sharpened proxy geometry $\mathbf{z}^*$ using a scaling factor $\beta_{\text{stab}}$ much higher than the $\beta$ we use in Eq.~\ref{eq:softmax_general}:
\begin{equation}\label{eq:stab_mean}
    \bar{z}_k = \frac{1}{N_e} \sum_{(i,j)\in E} z_{ij,k}
\end{equation}
\begin{equation}\label{eq:stab_softmax}
    z_{ij,k}^* = \frac{\exp(\beta_{\text{stab}} (z_{ij,k} - \bar{z}_k))}{\sum_{l=1}^3 \exp(\beta_{\text{stab}} (z_{ij,l} - \bar{z}_l))}
\end{equation}
This globally centered projection serves as a high-fidelity proxy of the final discrete geometry, operating on the assumption that design vectors will linearly diverge from the global mean under continuous binarization pressure. The moving displacement of this proxy design is evaluated and used as an auxiliary objective function $\mathcal{L}_{\text{stab}} = -x_{0}(T)$, which we term a \textit{stability pass}, in contrast to the ordinary objective evaluation with $\beta$, referred to as a \textit{performance pass}. During optimization, stability passes are interleaved with performance passes, and the resulting design variable updates are blended according to a predefined ratio. However, the aggressive sharpening induced by $\beta_{\text{stab}}$ can lead to severe vanishing gradients. To mitigate this, we adopt a Straight-Through Estimator (STE) to compute the gradient in a more reliable way \cite{bengio2013estimatingpropagatinggradientsstochastic}.
\begin{figure}
    \centering
    \includegraphics[width=0.9\linewidth]{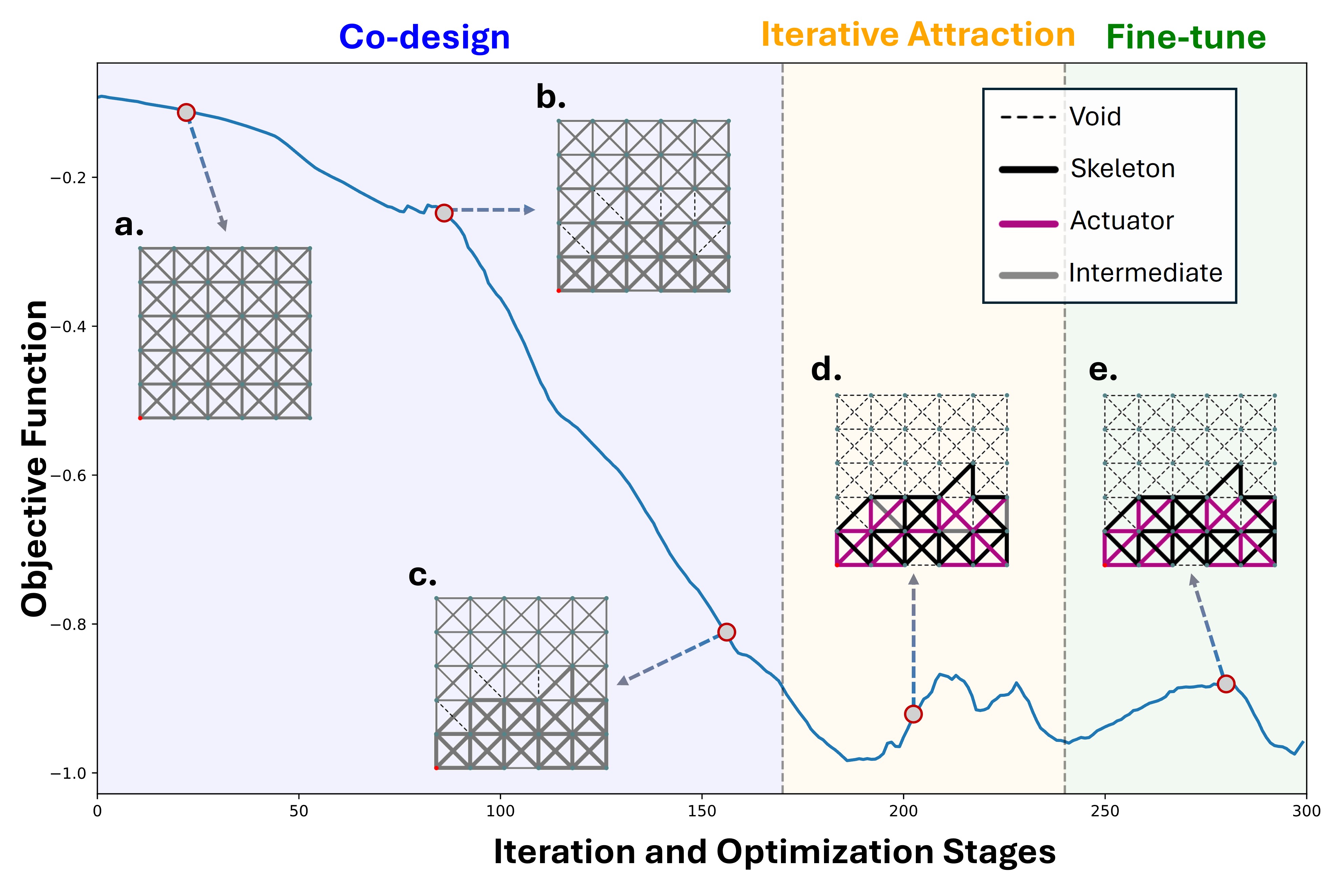}
    \caption{Optimization history of the co-design process corresponding to the design in Figure.~\ref{fig:init}b. (a–d) Intermediate designs at iterations 20, 70, 110, and 200, respectively, showing the evolution of the structure. (d) Structural fine-tuning during the design vector attraction stage. (e) Fully binarized and frozen structural layout in the control-only design stage, where only the MLP controller weights are fine-tuned.}
    \label{fig:sequence}
\end{figure}

To accelerate convergence during the later stages of optimization, where the continuous material design vectors $\mathbf{z}_{ij}$ exhibit reduced mobility due to smaller gradients (Fig. \ref{fig:manifold}d), we implement a iterative vector attraction phase (Fig. \ref{fig:sequence}). If the maximum value of a material state vector $\tilde{z}_{ij}$ exceeds a confidence threshold $\tau$, we deem the largest entry as the dominant state. The design vector is progressively pulled toward its dominant discrete material state using an additive nudge $\gamma$:
\begin{equation}\label{eq:attraction}
    z_{ij,c^{*}} \gets z_{ij,c^{*}} + \gamma, \quad z_{ij,k} \gets z_{ij,k} - \frac{\gamma}{N-1} \text{ for } k \neq c^{*},
\end{equation}
where $c^*$ is the index of the dominant material state.

Both the confidence threshold $\tau$ and the nudge strength $\gamma$ are linearly annealed over time. A hard discretization snap is enforced toward the end of the optimization process, after which the strictly binarized topology and material layout are frozen, and the neural controller parameters $\boldsymbol{\theta}$ are fine-tuned alone for the remaining iterations (Fig. \ref{fig:sequence}).

\begin{algorithm}
\caption{Multi-Stage Co-Design Optimization}
\label{alg:optimization}
\begin{algorithmic}[1]
\State \textbf{Input:} Initial design vectors $\mathbf{Z} \in \mathbb{R}^{N_e \times 3}$, MLP $\boldsymbol{\theta}$, Hyper-parameters.
\State \textbf{Output:} Discrete design $\mathbf{Z}^*$, fine-tuned controller $\boldsymbol{\theta}^*$.
\State Init. multipliers $\boldsymbol{\lambda} \gets \mathbf{0}$, penalties $\boldsymbol{\tau} \gets \tau_0$.

\For{$Iter = 1$ \textbf{to} $300$}
    \State Interpolate $T_{\text{grad}}$ and volume tolerance $\Delta_V$.
    \State Alternate passes $\in \{\textit{Controller, Performance, Stability}\}$. 

    \If{$Iter < 240$} \Comment{Phase: Co-design}
        \State Project $\mathbf{Z} \to \mathbf{\tilde{Z}}$ via softmax $\beta$ (Eq.~\ref{eq:softmax_general}).
        \State Simulate dynamics for $T$ steps (Eq.~\ref{eq:length}--\ref{coulomb}).
        \State Compute $\mathcal{L}_{\text{disp}} = -x_0(T)$ and constraints $v_k$.
        
        \If{\textit{Controller} pass}
            \State Update $\boldsymbol{\theta}$ via Adam w.r.t. $\mathcal{L}_{\text{disp}}$.
        \ElsIf{\textit{Performance} pass}
            \State Update $\mathbf{Z}$ via MMA w.r.t. AL obj. (Eq.~\ref{eq:opt_problem}).
        \ElsIf{\textit{Stability} pass}
            \State Project proxy $\mathbf{Z}^*$ via centered softmax $\beta_{\text{stab}}$.
            \State Compute $\mathcal{L}_{\text{stab}}$ \& update $\mathbf{Z}$ via STE + MMA.
        \EndIf
        
        \State Update AL params $\boldsymbol{\lambda}, \boldsymbol{\tau}$ using violations $v_k$.
        \If{$Iter \ge 170$} \Comment{Phase: Iterative Attraction}
            \State Apply iterative attraction(Eq.~\ref{eq:attraction}).
        \EndIf
        
    \ElsIf{$Iter == 240$} \Comment{Phase: Fine-tune}
        \State Snap $\mathbf{Z}$ to $\{0,1\}^3$ one-hot encodings.
        \State Freeze layout: $\mathbf{Z}^* \gets \mathbf{Z}$.
        
    \Else
        \State Simulate dynamics using frozen geometry $\mathbf{Z}^*$.
        \State Update only $\boldsymbol{\theta}$ via Adam w.r.t. $\mathcal{L}_{\text{disp}}$.
    \EndIf
\EndFor
\State \Return $\mathbf{Z}^*, \boldsymbol{\theta}^*$
\end{algorithmic}
\end{algorithm}

\subsubsection{Implementation Details}

\quad The following hyperparameters are chosen based on preliminary experiments. In this study, the physical simulation is evaluated for up to 300 design iterations. Each rollout integrates 8192 timesteps of $\Delta t=0.002$ s using a symplectic first-order Euler method in a differentiable simulation platform, Taichi, as described in Section 2.2. To preserve dynamic fidelity while accelerating convergence, gradients are initially calculated for only 2048 timesteps, and are gradually extended to a maximum of 4096 timesteps between iterations 80 and 150. The volume bound tolerance $\Delta_V$ is linearly widened from zero to $0.03$ between iterations 110 and 150. Iterative vector attraction begins at iteration 170, followed by a hard discretization snap at iteration 240. The neural controller is then fine-tuned on the strictly binarized topology for the remaining iterations (240–300). The complete pseudo-code implementation is shown in Algorithm \ref{alg:optimization}.

We select $N_{\text{CPG}} = 10$ CPG signals and a fundamental frequency of $\omega = 10$ following \cite{hu_difftaichi_2020}. Furthermore, we found that utilizing more than 10 CPG signals tends to desynchronize the robot's movement and harm overall locomotion stability. Similarly, to ensure the controller remains lightweight and stable during the highly non-convex co-design process, we adopt a hidden layer dimension of 32. The final MLP has layer widths of $(N_{CPG} + 4N_{m} + 2, D_{hidden}, N_{e})=(156,32,110)$.

The specific material properties assigned to the void, skeleton, and actuator states are summarized in Table \ref{tab:material_props}. Hyperparameters for the constraints, Augmented Lagrangian penalties, and optimization solver are listed in Table \ref{tab:hyperparameters}.

\begin{table}[h!]
\centering
\caption{Material properties assigned to the three discrete states.}
\label{tab:material_props}
\begin{tabular}{lccc}
\toprule
Material State & $\text{k}$ (N/m) & $\rho$ (g/m) & $a_{\max}$ \\
\midrule
Void ($V$) & $1 \times 10^{-7}$ & $1 \times 10^{-5}$ & $0.0$ \\
Skeleton ($S$) & $4 \times 10^{2}$ & $30.0$ & $0.0$ \\
Actuator ($A$) & $3 \times 10^{1}$ & $100.0$ & $0.35$ \\
\bottomrule
\end{tabular}
\end{table}

\begin{table}[h!]
\centering
\caption{Hyperparameters for simulation and optimization.}
\label{tab:hyperparameters}
\begin{tabular}{llc}
\toprule
Parameter & Symbol & Value \\
\midrule
Integration time step & $\Delta t$ & $0.002$ s \\
Total rollout steps & $T$ & $8192$ \\
Max topology volume fraction & $V^{\max}$ & $0.50$ \\
Min topology volume fraction & $V^{\min}$ & $0.50$ \\
Min actuator volume fraction & $V_{\text{act}}^{\min}$ & $0.20$ \\
Max actuator volume fraction & $V_{\text{act}}^{\max}$ & $0.22$ \\
AL initial penalty weight & $\tau_0$ & $0.3$ \\
AL penalty annealing factor & $a$ & $1.01$ \\
Performance pass inverse temp. & $\beta$ & $20.0$ \\
Stability pass inverse temp. & $\beta_{\text{stab}}$ & $500.0$ \\
STE inverse temp. & $\beta_{\text{STE}}$ & $20.0$ \\
\bottomrule
\end{tabular}
\end{table}

\section{Results and Discussion}
\subsection{Effect of Design Initialization}
\label{section31}
\begin{figure}
    \centering
    \includegraphics[width=0.9\linewidth]{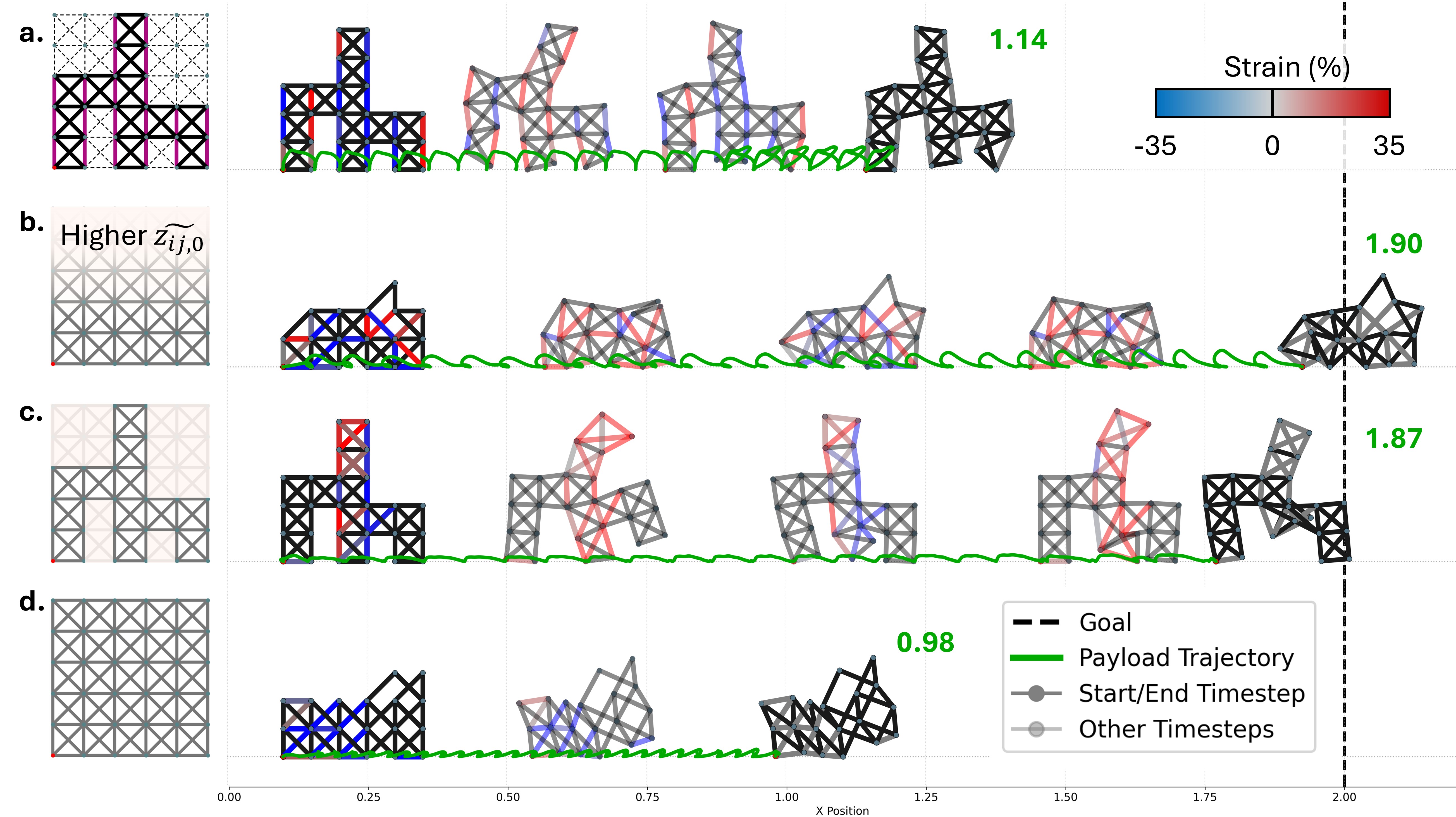}
    \caption{The baseline and optimization results with varying initial designs. (a) Vertical actuation layout baseline without optimization. (b)-(d) Optimization results from different initial heuristics: (b) Stability heuristic : top springs receive a higher void ratio. (c) Three-legged heuristic: shaded region receives a higher void ratio. (d) No heuristics.  The target position is marked by a vertical dashed line. Payload node's trajectories and final horizontal displacements of the robot are plotted in green, while instantaneous actuator strains are visualized with red denoting extension and blue denoting contraction.}
    \label{fig:init}
\end{figure}

We first investigate whether the proposed platform can effectively navigate the mixed-variable design space to discover optimal configurations of topology, material layout, and control. Furthermore, we verify whether co-design yields improved dynamic performance by comparing our method against a baseline that designs configuration and control separately, under identical volume fraction constraints.

We adapt the baseline design (Figure \ref{fig:init}a) from \cite{hu_difftaichi_2020}, which features a three-legged topology with vertically actuated active trusses and passive cross-braces to resist gravity. This configuration follows standard heuristics for walking robots, with three limbs to support an alternating gait. Additionally, it serves as the sequential design baseline where the pre-designed structural and material distributions are first fixed, then followed by the optimization of the neural network controller with Xavier-initialized weights. To provide a non-zero initial gradient, the payload node (bottom-left) is offset to $x=0.1$ m for all robots. The target positional vector is set to $(2, 0.1)$ m. 

The baseline robot learns an optimal control policy for its fixed material layout over 200 iterations, achieving a final horizontal displacement of 1.14 m.  However, the prevalent passive cross-bracing restricted the stroke magnitude of its limbs, causing inconsistent ground contact dynamics that destabilize its gait during later stages of the rollout, as indicated by the increasing disorderness in the payload trajectory.

Next, we compare the baseline to Design \ref{fig:init}b, which is fully co-designed using the proposed framework. To ensure a fair comparison, we constrained both robots to the same total topology and actuation volume budgets. Design \ref{fig:init}b is initialized using a gentle stability heuristic: trusses are assigned an initial ratio bias toward the void state proportional to their vertical elevation (y-height), with the topmost truss receiving a maximum ratio shift of $\Delta \tilde z_{ij,0} = +12\%$. The stability heuristic is aimed at preventing structural collapse, commonly caused by sparse, top-heavy geometries or large, unsupported active clusters. By promoting dense, low center of mass structures, this heuristic narrows the design space to favor solutions better suited for dynamic stability. The optimization trajectory of each edge design vector is visualized in Figure \ref{fig:manifold}, and the corresponding structural evolution of the entire robot is shown in Figure \ref{fig:sequence}. Upon initialization, all truss design vectors lie in intermediate states. As the optimization proceeds, most design vectors migrate rapidly and smoothly toward discrete states within 150 iterations under the Augmented Lagrangian binarization penalty (Figures \ref{fig:manifold}c). Volume fraction constraints encourage the solver to better allocate actuators and structural elements, leading to a significant decrease in the objective function and a corresponding improvement in the walking distance (Figure \ref{fig:sequence}a–c). Meanwhile, some trusses struggle to converge to a single material state (Figures \ref{fig:manifold}d and \ref{fig:sequence}c). These intermediate trusses are addressed in a second stage, where we introduce the progressive nudge of the iterative attraction phase to achieve binarization (Figure \ref{fig:sequence}d). During this process, those trusses gradually resolve the competition between performance and stability objectives, indicative in their zigzagging trajectories (Figure \ref{fig:manifold}e). Ultimately, the design achieves full convergence before the hard binarization snap (Figure \ref{fig:sequence}e), with additional iterations devoted to further control optimization. The final converged morphology for Design \ref{fig:init}b adopts a propulsive leaping gait, relying primarily on a massive back leg for thrust while utilizing a smaller front leg for traction. This co-designed robot achieves a final displacement of 1.90 m, outperforming the sequential baseline by 66\%.

However, the notable performance improvement demonstrated by \ref{fig:init}b could be simply attributed to a superior initial heuristic compared with the baseline. To test this possibility, we introduce a different initial solution in Design \ref{fig:init}c that mirrors the three-legged configuration of the baseline, allowing us to examine whether co-design can consistently improve locomotion performance across alternative initializations. Here, trusses that are void in the baseline (Design \ref{fig:init}a) are initialized with a strong bias toward the void state, while all remaining active and skeletal trusses start with completely uniform and equal ratios for each state. During optimization, Design \ref{fig:init}c concentrates its actuators within the middle leg (Figure \ref{fig:init}c). This allocation overcomes the restricted stroke observed in the baseline, enabling sweeping motions and lengthened strides to achieve a final displacement of 1.87 m. This result falls slightly short of Design \ref{fig:init}b, indicating that the performance indeed depends on initialization, as expected for such a nonconvex optimization problem. However, it still substantially outperforms the baseline that shares the same three-legged initial configuration. This result highlights the advantage of co-design over separated design.

Finally, we verify that even for the most challenging scenario, a heuristics-free initialization (Figure \ref{fig:init}d), the proposed platform can still find a viable solution. Here, optimization begins with completely uniform material ratios for the void, skeleton, and actuator states across all trusses. The algorithm receives no initial guidance and is forced to search for a structurally valid and dynamically favorable solution from an unrestricted design space. Despite this lack of guidance, the co-design platform is still able to discover a viable design that achieves a travel distance of 0.98m. The resulting Design \ref{fig:init}d adopts a hopping gait: using its front leg as support and frugally allocating the highly compliant muscle near the back leg payload for motion. Although the heuristic-free optimization succeeds in finding a valid locomotion strategy, it still underperforms relative to Designs \ref{fig:init}b and \ref{fig:init}c, as we intend to choose a challenging initial solution to test optimization sensitivity. Future gradient-based co-design approaches could be combined with global search methods, such as genetic algorithms, to seed the solver with strong initial solutions. Interestingly, Design \ref{fig:init}d automatically discovered structural features similar to the stability-heuristics Design \ref{fig:init}b despite receiving no guidance. Common emergent features include disconnecting specific bottom trusses for larger strides, and concentrating topology to the lower half for a low center of gravity.

\subsection{Individual and Collective Contribution of Design Entities}
\label{section32}
\begin{figure}
    \centering
    \includegraphics[width=\linewidth]{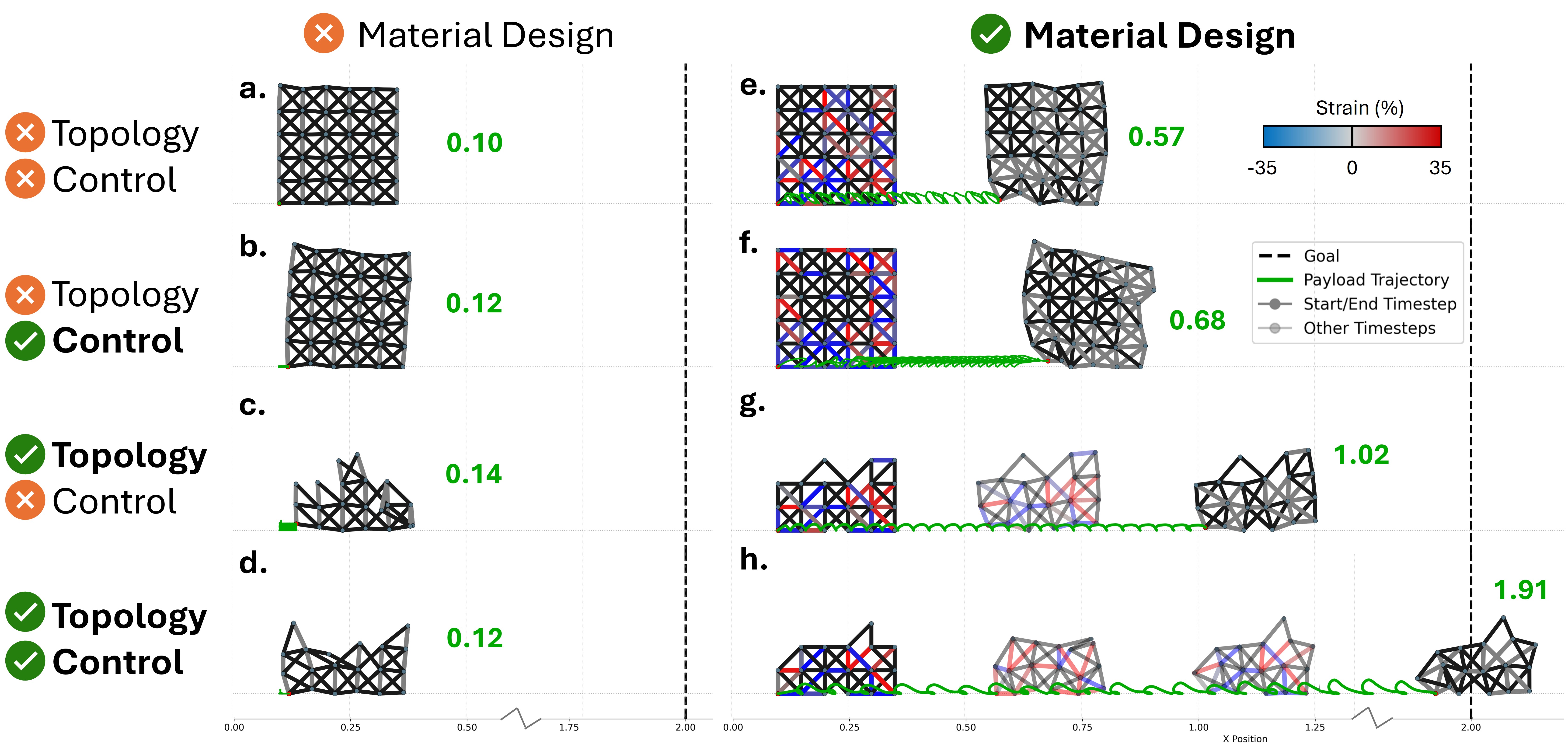}
    \caption{Combinatorial ablation study results. Robots are optimized from an identical initialization. A \textcolor{RedOrange}{cross mark} indicates a disabled design entity, while a \textcolor{ForestGreen}{tick mark} indicates indicates an enabled entity. }
    \label{fig:ablation}
\end{figure}

\begin{figure}
    \centering
    \includegraphics[width=0.7\linewidth]{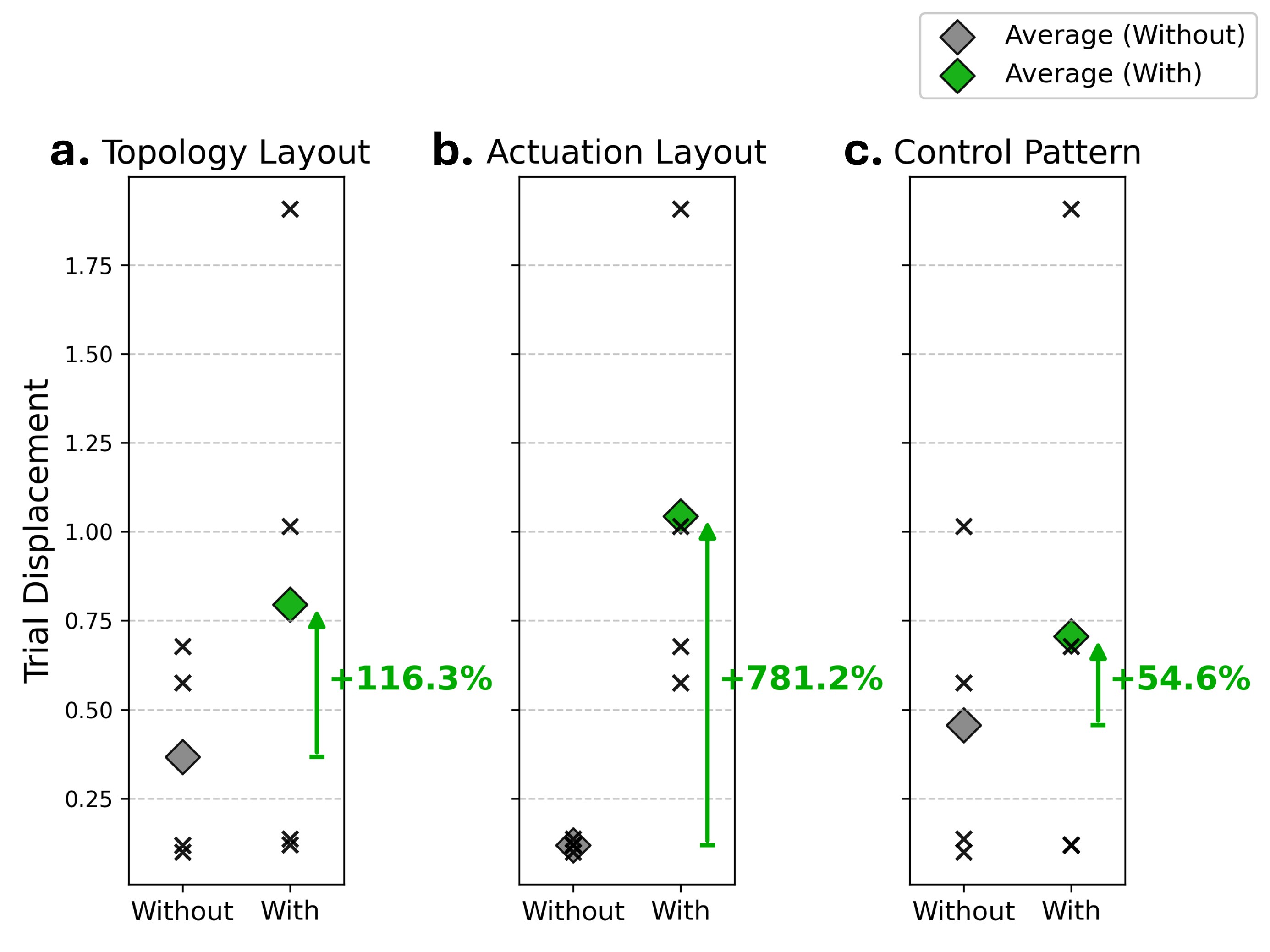}
    \caption{Quatitative ablation results. Grouped by design entities, the average with and without performance for each entity is compared. Actuation layout has the largest contribution to improved performance, while control pattern has the smallest. }
    \label{fig:contribution}
\end{figure}
We continue to leverage our platform to investigate the individual and collective contributions of design entities to locomotive performance. We conduct ablation studies by selectively disabling specific entities, allowing us to isolate their individual effects or examine pairwise combinations. We then compare these ablated trials with the fully co-designed case to quantify the benefits of jointly optimizing all three design entities.

By combinatorially considering the enabled/disabled design of the three design entities, we obtain eight possible design schemes or trials (Figure \ref{fig:ablation}). Specifically, trials with the control design disabled (Designs \ref{fig:ablation}a, c, e, and g) operate using frozen MLP weights from Xavier Initialization. For no-topology-design trials (Designs \ref{fig:ablation}a, b, e, and f), a default fully-filled topology is used, while a uniform distribution of vertical actuators as used the default actuation layout for no-material-design trials (Designs \ref{fig:ablation}a, b, c, and d). The same initialization in Figure \ref{fig:init}b is applied to all trials that enable topology design (Designs \ref{fig:ablation}c, d, g, and h). To ensure the topology remains frozen while materials can still be redistributed,  Trials \ref{fig:ablation}e and f are subject to a topology volume fraction of $V^{\min}=0.90$, which is only appropriately strict to prevent sharp design vector motions at the simplex edges (as seen in Figure \ref{fig:manifold}c). Conversely, to freeze the material layout while permitting topological changes in Trials \ref{fig:ablation}c and d, we subtract a small constant from the non-dominant solid entries of the design vector before every softmax projection, whereas the void entries are left unaltered to enable topological design.

The quantitative results are summarized in Figure \ref{fig:contribution}. Actuation layout demonstrates the largest individual contribution to performance improvement, increasing the average displacement by 781\% compared to the "without" conditions (Figure \ref{fig:contribution}b). Admittedly, the default actuator distribution imposes stringent requirements on compatible topologies, which must mechanically facilitate high-stroke motions for traversal. Yet, actively co-designing the actuation layout consistently yields substantial performance improvements regardless of the underlying topology (comparing Trials \ref{fig:ablation}a–d to \ref{fig:ablation}e–h), highlighting its high adaptability to varied initial conditions and its values in robotic design. Topological design produces an improvement of 116\% (Figure \ref{fig:contribution}a), while weak on its own, it is a powerful multiplier for the positive impact of material design. By comparing trials optimized with and without topology while holding the other two design entities constant (Trial \ref{fig:ablation}e vs. g, and \ref{fig:ablation}f vs. h), we observe that topological optimization consistently increases traversal distance, even on top of the already substantial gains achieved by actuation design alone.

Surprisingly, control pattern design demonstrates a small individual contribution and occasionally impacts performance negatively (Trial \ref{fig:ablation}c vs. d). This behavior likely occurs because the topology and material distribution themselves are suboptimal and provide limited mobility, leaving little room for further improvement through control optimization alone. The degradation may also arise because the severe non-linearities introduced by ground contacts pollute the gradient propagation through time; when coupled with a sub-optimal physical structure, the resulting small loss, and consequently small gradient amplitude, muddies the directional information provided to the Adam optimizer. However, when combined with material layout optimization, control consistently enhances traversal, as observed in Trials \ref{fig:ablation}e vs. f and \ref{fig:ablation}g vs. h. Most notably, when we further incorporate the topology entity, co-evolving control alongside all other entities, travel distance increases by 87\% compared to the trial without control optimization (from 1.02 m in Trial \ref{fig:ablation}g to 1.91 m in Trial \ref{fig:ablation}h). This result demonstrates that the benefits of jointly optimizing topology, material distribution, and control exceed those achievable by optimizing the entities in isolation.

Additionally, we observe that in trials producing substantial displacement (Designs \ref{fig:ablation}e–h), the resultant material distributions and locomotion strategies are topology-dependent. For example, the fully-filled topologies in Trials \ref{fig:ablation}e and f both allocate actuators heavily to the right and bottom edges of the robot to maximize traction. Conversely, the freely evolved topologies in Trials \ref{fig:ablation}g and h allocate actuators to both the rear and front legs while independently discovering an identical central motif: inserting a horizontal actuator directly between passive cross-braces for enhanced coordination between the front and rear legs. This phenomenon mirrors convergent evolution in nature, where identical musculoskeletal motifs emerge across species with similar body plans. This further highlights the importance of simultaneously optimizing all design entities. Discovering a highly optimized topology without its uniquely compatible actuation layout (Trial \ref{fig:ablation}d vs. h), or vice versa (Trial \ref{fig:ablation}f vs. h), leads to substantially weaker performance. This outcome indicates a strong coupling between material and topology, emphasizing that the two must be jointly designed to enable optimal performance.

\subsection{Generality to Other Tasks and Environments}
\label{section33}
\begin{figure}
    \centering
    \includegraphics[width=0.9\linewidth]{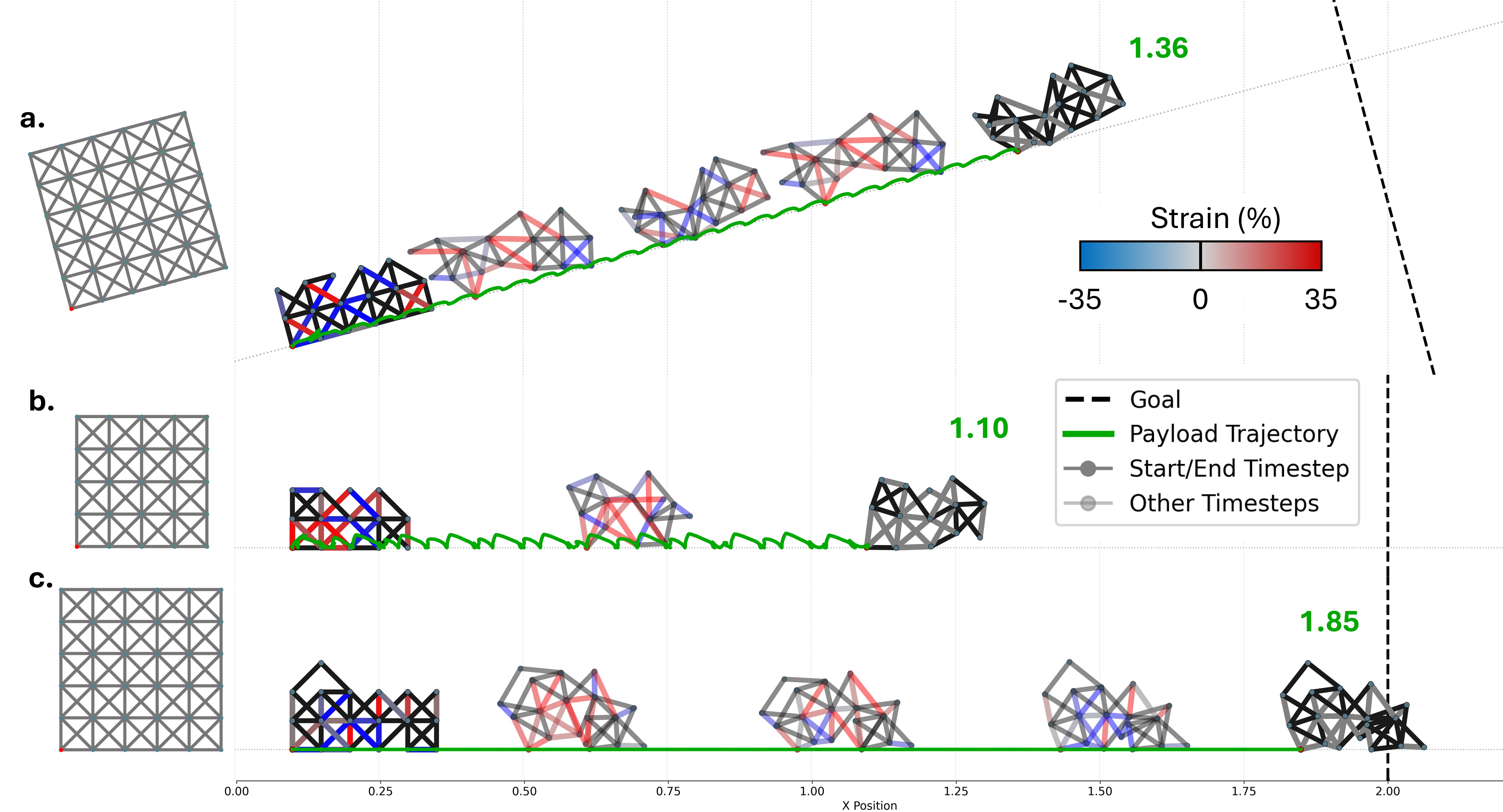}
    \caption{Alternative tasks and environments: (a) upward incline; (b) $5\times5$ nodes design space; and (c) slippery ground contact. }
    \label{fig:cases}
\end{figure}

Finally, we apply the proposed co-design framework to several additional cases with varying optimization objectives, design space, and environmental parameters to demonstrate its flexibility. All cases begin with the stability heuristics described in Section \ref{section31}. 

In design \ref{fig:cases}a, we place the locomotion goal on top of a 15-degree incline. Although sharing the same initialization as Design \ref{fig:init}b, the final robot morphology changed in adapting to this new target function. Compared to Design \ref{fig:init}b, the final design \ref{fig:cases}a evolved to an even slimmer structure to further prioritize a low center of mass, crucial for stabilizing against the toppling forces of the inclined ground boundary. Dynamically, it adopts an arching motion to gain altitude as well as forward distance, that is a combination of hopping (Design \ref{fig:init}c) and crawling (Design \ref{fig:init}d). Additionally, it slightly undercuts the lower volume bound for topology and actuation in exchange for having to carry less weight upwards, achieving a horizontal distance of 1.36 m and vertical elevation of 0.35 m. 

For Design \ref{fig:cases}b, we perform optimization on a smaller design space of $5\times5$ nodal grid, rather than a $6\times6$ nodal grid in all other cases. This reduction in available material resources makes the allocation of materials and topology more critical. Our co-design platform reliably generalizes to this design space and automatically rediscovered the leg division strategy observed in designs Design \ref{fig:init}b and d, where the bottom trusses are broken to form a rear and front leg. It used a front-leg assisted leaping gait, primarily relied upon the rear actuators, and achieved a traversal distance of 1.10 m upon final evaluation, comparable to that of Design \ref{fig:init}d. 

Our proposed framework is also flexible enough to handle varying physical and environmental parameters. To demonstrate this capability, we apply it to a locomotion design problem under different ground friction conditions, which strongly influence contact mechanics and locomotion dynamics. Specifically, Design \ref{fig:cases}c adopts the Coulomb friction contact model specified in Section \ref{sec:evaluation}, differentiating itself from the infinite friction model used in all other designs. The coefficient of dynamic friction is chosen as $\mu=2.5$ \cite{roth_frictional_1943}. Under this setting, it exhibits a consistent morphological adaptation to the new physical environment, adopting a locomotion strategy unobserved in previous designs. Instead of dynamically lifting the payload and repositioning it forward with every gait, as observed in Designs \ref{fig:ablation}g and h with similar topology, it keeps the heavy payload with the ground to provide anchorage against skidding. Interestingly, during the first 0.5 meters, the robot folds its front leg inward, increasing the contact area and effectively using it as a friction pad. This behavior represents a beneficial morphological transformation that leverages the coupled effects of environmental physics, topology, material layout, and actuation control. 

\section{Conclusion}
\label{conclusion}
In this work, we have developed a co-design framework to automatically identify the optimal combination of topological layout, distribution of passive and active materials, and control policy of a discrete truss-lattice robot, maximizing its performance across a range of dynamic locomotive tasks. Within this framework, we construct a unified design representation that links mixed variable structural and material design with a spatiotemporal neural network controller. This formulation allows different design entities to be coupled within a differentiable physics simulator that efficiently evaluates system dynamics and provides gradient information, even under frictional contacts and varying environments. Unlike most existing methods that treat these design entities separately, our approach optimizes them jointly within a constrained optimization framework that accounts for structural stability and physical feasibility, providing greater flexibility to achieve improved performance.

Through a variety of design cases, we demonstrate that the proposed framework reliably discovers high-performance solutions even without heuristic guidance. With only simple initialization, it can outperform carefully crafted baseline designs developed through sequential design processes. Across tasks, the framework identifies diverse material allocations and locomotion strategies for the same traversal objective, such as leaping, sweeping, and hopping gaits, indicating that co-design expands the accessible morphological design space and enables not only higher performance solutions but also a broader range of viable designs. The framework also adapts to different design requirements and environmental conditions, automatically generating robots whose structures, material distributions, and control policies are tailored to specific tasks to produce effective motion patterns. These results highlight both the adaptability of the proposed method and the importance of co-design for enabling niche robotic systems that accommodate the specific requirements of customized applications.

Further, this study uncovers the contributions of structural, material, and control design entities in robotic locomotion. We found that material layout plays the most critical role, as redistributing actuation within a given configuration can effectively reshape robot dynamics and achieve strong performance even when the structure and control are suboptimal. In contrast, control alone cannot compensate for limitations in structure or actuation and is largely bounded by the underlying structural and material design. More importantly, the interaction among structure, materials, and control produces performance gains that exceed the contributions of any single component. We also observe convergent structural motifs emerging across designs with similar objectives, suggesting an underlying form-function relationship that arises from the synergy among these design entities. These insights further highlight the importance and value of simultaneous co-design realized in the proposed framework.

There are several limitations in this work that motivate future research. First, as expected for highly nonconvex optimization problems, the final performance remains sensitive to initialization. A promising direction is to combine the proposed gradient-based search with global optimization methods, such as evolutionary algorithms \cite{gu_optimization_2025, strgar_accelerated_2025}, to improve exploration. Second, as an exploratory study, we employ a simplified simulator. Although the corresponding physical system has recently been realized with good agreement to simulation \cite{strgar_evolution_2024}, incorporating richer physical effects such as bending and material nonlinearities would further improve fidelity. In addition, while we currently optimize the neural controller using gradient-based methods, integrating reinforcement learning strategies may provide greater generalization. Finally, accounting for manufacturing variability and environmental uncertainty will be important for enabling more robust robotic designs.

We envision that the proposed framework can be broadly adopted to establish design principles and develop computational tools for a wide range of robotic systems. By customizing each component, the approach is generalizable and applicable to diverse materials and stimuli. We anticipate that it will accelerate the development of robotic material systems with various applications, including bio-hybrid robots \cite{kinjo2024biohybrid, sun2020biohybrid}, mechanical computing systems\cite{liang2024physical, lee2022mechanical}, and more.

\section*{Acknowledgments}
We acknowledge support from the Department of Mechanical Engineering at Carnegie Mellon University.

\bibliographystyle{unsrt}
\bibliography{asme2e}

\end{document}